\documentclass{bmvc2k}
\usepackage{graphicx}
\usepackage{comment}
\usepackage{color}
\usepackage{subcaption}
\usepackage{amsmath,amsfonts}
\usepackage{amssymb,amsopn}
\usepackage{bm} % bold symbol

\usepackage{amssymb}
\usepackage{amsmath,amsfonts}
\usepackage{amsthm,amsopn}

\usepackage{bm} % bold symbol

\newcommand{\ProbOpr}[1]{\mathbb{#1}}
\newcommand{\expect}[2]{%
\ifthenelse{\equal{#2}{}}{\ProbOpr{E}_{#1}}
{\ifthenelse{\equal{#1}{}}{\ProbOpr{E}\left[#2\right]}{\ProbOpr{E}_{#1}\left[#2\right]}}} % Expectation: syntax: E{1}{2} = E_1[2], E{}{2}=E[2], E{1}{} = E_1
\newcommand{\var}[2]{%
\ifthenelse{\equal{#2}{}}{\ProbOpr{VAR}_{#1}}
{\ifthenelse{\equal{#1}{}}{\ProbOpr{VAR}\left[#2\right]}{\ProbOpr{VAR}_{#1}\left[#2\right]}}} % Expectation: syntax: V{1}{2} = V_1[2], V{}{2}=V[2], V{1}{} = V_1
\DeclareMathOperator{\argmin}{arg\,min}

\newcommand{\eat}[1]{} 

\title{Online Action Detection in Streaming Videos with Time Buffers}

\addauthor{Bowen Zhang}{zhan734@usc.edu}{1}
\addauthor{Hao Chen}{hxen@amazon.com}{2}
\addauthor{Meng Wang}{mengw@amazon.com}{2}
\addauthor{Yuanjun Xiong}{yjxiong@amazon.com}{2}

\addinstitution{
 U. of Southern California\\
 Los Angeles, USA
}
\addinstitution{
 Amazon\\
 Seattle, USA
}

\runninghead{Zhang et. al,}{Online Action Detection in Streaming Videos}

\begin{document}

\maketitle

\begin{abstract}
We formulate the problem of online temporal action detection in live streaming videos, acknowledging one  important property of live streaming videos that there is normally a broadcast delay between the latest captured frame and the actual frame viewed by the audience.
The standard setting of the online action detection task requires immediate prediction after a new frame is captured. We illustrate that its lack of consideration of the delay is imposing unnecessary constraints on the models and thus not suitable for this problem.
We propose to adopt the problem setting that allows models to make use of the small ``buffer time'' incurred by the delay in live streaming videos.
We design an action start and end detection framework for this \textbf{online with buffer} setting with two major components: flattened I3D and window-based suppression.
Experiments on three standard temporal action detection benchmarks under the proposed setting demonstrate the effectiveness of the proposed framework.
We show that by having a suitable problem setting for this problem with wide-applications, we can achieve much better detection accuracy than off-the-shelf online action detection models.

\end{abstract}

%%%%%%%%% BODY TEXT
\section{Introduction}

\newcommand{\buffer}{buffer}

 Video temporal action detection is to predict what actions a video contains as well as to localize where the actions are in the video timeline. The problem is mainly studied in the offline setting today, in which the full video is accessible for use in the detection~\cite{zhao2017temporal,shou2016temporal,shou2017cdc}. In many of today's use cases, however, it is needed to detect actions in a {\it real time online} manner without the access of the full videos. One important application of online temporal action detection is to detect the actions or events of interests in real-time in {\it live streaming videos}. For instance, by detecting the moment that players jump the ball in a live broadcast of an NBA game, we know that the match starts; Another example can be to send an in-time red flag for a live webcast, if not safe for work (NSFW) contents such as nudity or violent behaviors are detected. With the rapidly growing number of diverse live streaming contents on the internet and media, there is an increasing need today to have some intelligent approaches to detect the start and the end of actions or events in live streaming videos.

The strict online action start and end detection setting requires the model to make decision immediately when it observe a new frame and not to make any change to its previous predictions. These two requirements make the problem inherently challenging. First, the ``no future information'' requirement makes the model lose half of the temporal context information for decision making. Second, since the predictions are all made in real-time in the strict online manner, there is no way to correct the wrong predictions in previous frames once the decision is made. An example of failure cases can be that the model continuously makes positive predictions while many of them are actually false alarms. Current online detection approaches~\cite{shou2018online,gao2019startnet} tackle these problems by either enlarging the training set via generating hard-negative features using generative adversarial networks, or using policy gradient to encourage the model to filter out hard negative frames. These approaches are effective to some extent but not overall satisfactory.
It also makes us question whether this strict setting is necessary for all applications.

 \begin{figure}
    \centering
    \vspace{-0.05in}
    \includegraphics[width=0.9\textwidth]{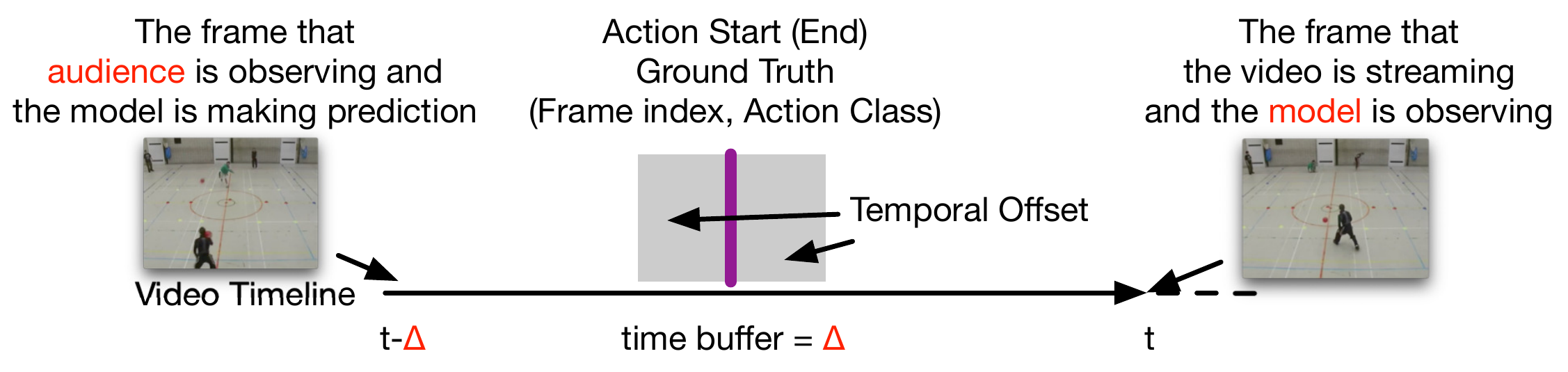}
    \vspace{-0.1in}
    \caption{Illustration of time buffer and the action start (end) detection for online videos. The video is played at frame $t$, and the audience is watching frame $t-\Delta$, due to the time delay. $\Delta$ is the time buffer. The model can take frames till $t$ as inputs, and make prediction on frame $t-\Delta$. The prediction is considered as a true positive only when the following two conditions are satisfied: (1) the frame index that is predicted as an action start/end is within the temporal offset of the ground truth, and (2) the action class is predicted correctly.
    }
    \label{fig:illustration}
    \vspace{-0.2in}
\end{figure}

We argue that the strict online setting of action start and end detection is unnecessary in a number of live streaming video use cases.
Unlike use cases for autonomous driving cars or robot systems in which complete real-time is pursued, in live streaming broadcast, e.g., YouTube, BBC, ESPN, CCTV Live, etc., it is common that the contents watched by audiences are slightly delayed from the original live videos. The short delay exists due to a variety of factors, such as the intentional delay designed to prevent profanity, bloopers, nudity, or other undesirable material from making it to air, including more mundane problems like technical malfunctions, known as the broadcast delay~\cite{broadcastdelay}; delays due to the network transmission and overhead; or delays made for the insertion of advertisement, and so on~\cite{livetv}. The delay can vary from several milliseconds to tens of seconds, depending on different cases. Since an online detection model can take the original live video as the input but only needs to make predictions before the video is received by the audiences, such delays relax the strict online detection setting and provide opportunities for the model to use a limited amount of future information in decision making. Because none of today's online temporal action detection approaches take such short time delays into consideration, those solutions may be sub-optimal for the real-life live streaming video use cases.

In this paper, we rethink the online temporal action detection problem for the live streaming video settings. We show that the relaxation of the strict online prediction setting with the time delay is more practical with this problem. The time delay, though short, can not only provide crucial near-term future information to support the prediction of the types of actions, but also help reduce the errors in predicting action boundaries by viewing more frames in advance and selecting the one with the highest confidence in a time window. Specifically in our relaxed problem setting, the model is able to utilize a limited amount of frames ahead of what audiences are watching, denoted as the {\it time buffer}, to predict whether there is an action start/end in the frame that the audiences are receiving at that moment. Fig.~\ref{fig:illustration} is an illustration of the {\it time buffer}. The model can observe frames till time $t$, and only need to make prediction on time $t-\Delta$, where $\Delta$ is the time buffer. Through this work, we aim to answer the following questions: (1) whether the time buffer is helpful for online action start and end detection; (2) how much performance improvement can we gain from the time buffer; (3) how the time buffer should be used to improve the accuracy of the detection. Our main contributions are:

\vspace{-0.1in}
\begin{itemize}
    \item We rethink the action start and end detection problem for live streaming videos by taking the {\it time delays} into consideration, which is currently ignored in literature;
    \vspace{-0.1in}
    \item We propose {\it buffer based flattened I3D (BF-I3D)}, a new paradigm of using time delays to predict the start and end of actions of interest in the live stream;
    \vspace{-0.1in}
    \item We conduct ablation studies on multiple widely used activity datasets. We further investigate these datasets to understand the applicability of them in evaluating the online action start and end detection task.
\end{itemize}

Our work is the first to consider a practial setting of online action start end detection with time buffer for live streaming broadcast and provide  corresponding solutions to the problem. Results show that even with a tiny time delay (e.g., less than 3 seconds), our approach can significantly increase the accuracy in the action start and end prediction. Our results establish the state-of-the-art of this problem with a $13.8\%$ improvement in detection mAP on THUMOS'14 dataset. We further study the effectiveness of today's activity datasets in bench-marking the action start and end detection approaches and provide insights on future dataset curation.

\section{Related Work}
\label{related}

\textbf{Action Recognition} aims to recognize human actions in videos~\cite{wang2013action,wang2011action,laptev2005space,wang2016temporal,simonyan2014two}.
Traditional methods for action recognition emphasize extracting hand-crafted features that capture space and time information~\cite{laptev2005space,dalal2005histograms,wang2011action,wang2013action}.
In the past years, there have been a large body of works utilizing deep learning to solve this topic, including 2D~\cite{karpathy2014large,simonyan2014two,feichtenhofer2016convolutional} and 3D CNN based methods~\cite{tran2015learning,carreira2017quo,feichtenhofer2019slowfast,wang2018appearance}, and recurrent networks~\cite{wu2015modeling,jiang2017exploiting}.
Among them, the 3D CNN based approaches are most related to this work.
3D CNN is first proposed in~\cite{tran2015learning} to tackle action recognition problem. Later works introduce large scale pretraining on YouTube videos~\cite{abu2016youtube} and transfer learning from ImageNet~\cite{carreira2017quo} to enhance the capacity of neural networks.
In~\cite{feichtenhofer2019slowfast}, the 3D convolution operation is further decomposed into two consecutive convolution operations specializing in the spatial and temporal axes.
In this work, an I3D model~\cite{carreira2017quo} pretrained on the Kinetics dataset~\cite{carreira2017quo} is used as a sliding window classifier for online action detection.

\noindent
\textbf{Temporal Localization for Offline Video.}
Temporal action localization aims at detecting the start and end time of action instances in an untrimmed video.
THUMOS14~\cite{THUMOS14}, ActivityNet~\cite{caba2015activitynet}, and HACS~\cite{zhao2019hacs} datasets are designed for this task.
Representative methods for this task include multi-stage CNN-based approaches~\cite{shou2016temporal,shou2017cdc,zhao2017temporal} and recurrent neural network based approaches~\cite{lin2017temporal}.
In~\cite{shou2017cdc}, temporal convolution is introduced to refine the detected action instance boundaries.
In~\cite{zhao2017temporal}, a learning based proposal generation method is introduced and has led to a series of works on learning better action proposals~\cite{lin2018bsn,lin2019bmn}.
In~\cite{zeng2019graph}, graph convolutional neural networks are used to extract the contextual relationships between action instances and help temporal action localization.

\noindent
\textbf{Online Detection of Human Action.}
Recently, online temporal action detection has drawn multiple attentions.
In~\cite{shou2018online}, the authors propose the task of online detection of action start and define the evaluation metrics for this task.
In~\cite{xu2019temporal}, a new encoder-decoder architecture is introduced to further improve the detection accuracy.
Our work differs from these works in that we recognize the realistic value of buffer time which is common in broadcasting system and propose a way to utilize it to achieve superior performance.

\begin{figure}
  \centering
  \includegraphics[width=\linewidth]{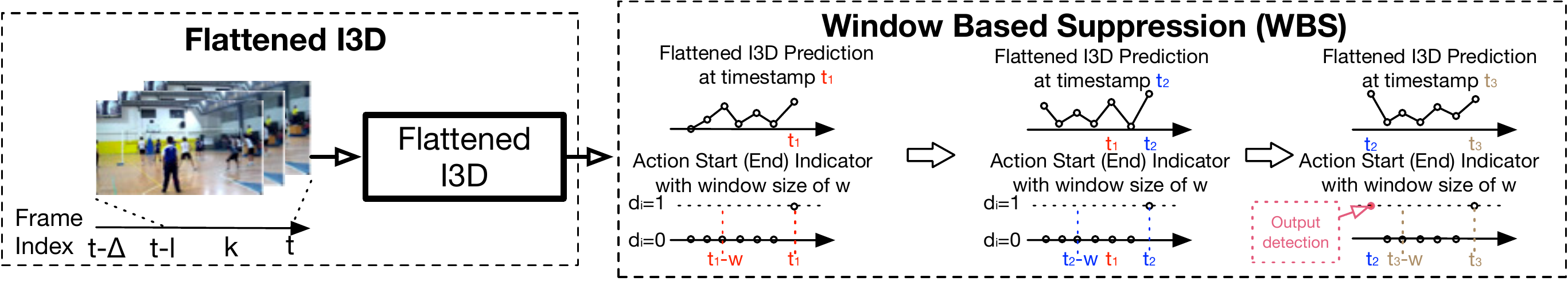}
    \caption{The overall framework of the buffer based flattened I3D (BF-I3D). BF-I3D contains Flattened I3D Module and WBS Module. The Flattened I3D takes frames from $t-l$ to $t$ as the input, and outputs the prediction for frame $k$. Predictions are then aggregated by the WBS with window size $w$ to update action start (end) indicators. The final prediction is outputted once the frame moves out of the window.}
\label{fig:architecture}
\end{figure}

\section{Approach}
\label{approach}

In this section, we first introduce the new online setting for live streaming videos, and formulate the problem of online action start and end detection with time delays. Then we introduce a new paradigm - a buffer based flattened I3D (BF-I3D) solution, which can leverage the information from time buffers in online action detection.

\subsection{Problem Formulation}
\label{buffer_setting}

Due to the existence of the time delays in the live streaming videos, the frame that audiences receive is lagging behind the current frame in the original video. For simplicity, in the rest of the paper, we denote the stream of frames that audiences receive as the {\it audience stream}, and the stream of the frames from the original video stream as the {\it source stream}.

Fig.~\ref{fig:illustration} illustrates the time buffer and the action start (end) detection for live streaming videos. Assume the video contains $n$ frames. We denote frames as $\{x_i\}_{i=1}^n$, and the time delay is $\Delta$, which is usually as small as a few seconds. Then we have a time buffer $\Delta$ between the source stream and the audience stream, i.e., assume the source steam is now at frame $t$, $t \in \{1,2,...n\}$, the audience is observing frame $t-\Delta$. The model can observe all frames till $t$, i.e., frames $\{x_i\}_{i=1}^{t}$, and only needs to make prediction for the $t-\Delta$ frame for the audience. The prediction includes whether the frame is an action start or end, and which action category the frame contains.
Note that under the time buffer setting, the audiences can still observe the streaming video without any pause. The audiences experience $\Delta$ frames delay from the beginning of the video streaming.

\subsection{Online Detection Model Framework}
\label{detection_framework}

We propose a {\it buffer based flattened I3D (BF-I3D)} approach for online detection of the action start and end in live streaming videos. BF-I3D is a sliding window based approach that takes a sequence of frames as input and outputs the decision of action start or end for one specific frame. Fig.~\ref{fig:architecture} illustrates the overall framework of the BF-I3D, which is mainly composed of two modules: the {\it Flattened I3D module} and the {\it window-based suppression (WBS) module}.

Specifically, assume the source stream is at frame $t$, the Flattened I3D module takes $l$ frames in the past as input, i.e., $\{x_{t-l}, \ldots, x_{t}\}$, and outputs a $C+1$ dimension probability vector $\bold{p}_k$ for the $k^\text{th}$ frame, where $C+1$ is the number of action classes plus the background:
\begin{small}
\begin{align}
    \bold{p}_k = f(\{x_{j}\}_{j=t-l}^{t}).
\end{align}
\end{small}
$f$ represents the Flattened I3D model. Each item $p_k^c$ in $\bold{p}_k$ represents the probability of the $k^\text{th}$ frame
that contains the start (or end) of the action category $c$, for $c \in \{1,2,3,...C\}$, or the probability to be the background if $c=C+1$. We restrict $t-\Delta \leq k \leq t$, so that the prediction is always ahead of the frame that the audience is currently receiving (i.e., the frame $t-\Delta$). It is worth to note that here time buffer enables the Flattened I3D to use short ``future'' information (i.e., frame $k+1$ to $t$) in predicting frame $k$.

Next, we generate the final prediction for the frame $t-\Delta$ with the WBS module.
For each frame $i$, we denote an indicator $\bold{d}_i$ that represents whether frame $i$ is an action start (end). $\bold{d}_i$ is a $C+1$ dimension vector, with each dimension as a binary number, i.e., $d_i^c = 1$ represents that frame $i$ is a start (end) of the action $c$, and $d_i^c = 0$ represents that frame $i$ is not.
Assume that the window size of the WBS module is $w = k - (t-\Delta)$, then the final prediction of frame $t-\Delta$, denoted as $\bold{y}_{t-\Delta}$, will be made only when the predictions from Flattened I3D module are made till frame $k$, while frame $t-\Delta$ is going to move out of the window of WBS. The WBS module takes the predictions $\bold{p}_i$ as inputs, and updates $\bold{d}_i$, $\forall i \in [t-\Delta, k]$:
\begin{small}
\begin{align}
    [\bold{d}_{t-\Delta}, ...,\bold{d}_{k}] = \textsc{wbs}([\bold{d}_{t-\Delta}, ..., \bold{d}_{k-1}], [\bold{p}_{t-\Delta}, ... , \bold{p}_{k}]).
\end{align}
\end{small}
Details of the WBS is referred to the following section. The final output at frame $t-\Delta$ is:
\begin{small}
\begin{align}
    \bold{y}_{t-\Delta} = \bold{p}_{t-\Delta} \cdot \bold{d}_{t-\Delta},
\end{align}
\end{small}
i.e., the time buffer enables to generate final prediction $\bold{y}_{t-\Delta}$ at $t-\Delta$ based on not only frames prior to it but also its $k$ succeeding frames.

The online detection without using the buffer is a special case in our formulation, with $\Delta = 0$. In that case, $t - \Delta = k = t$, i.e., the prediction is always at the current frame.

\begin{figure}
    \centering
    \includegraphics[width=0.8\textwidth]{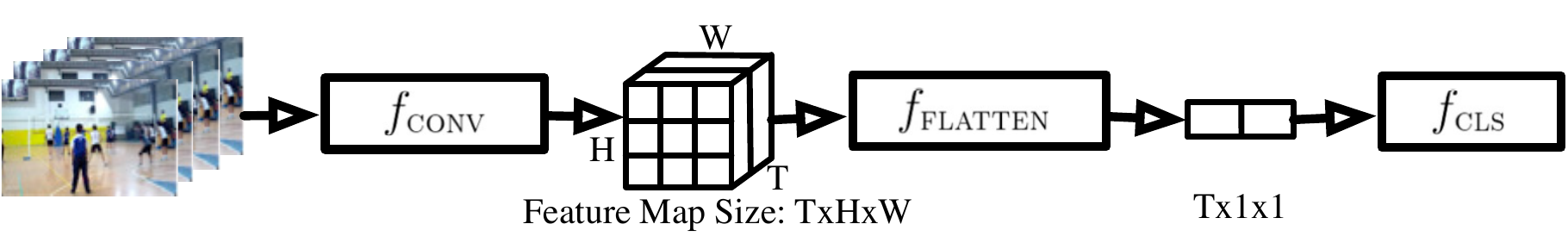}
    \caption{Structure of Flattened I3D.  $T$, $H$, and $W$ represent the temporal dimension, height, and width of the feature map (We omit the channel dimension here for simplicity). }
    \label{fig:flatten_structure}
\end{figure}

\paragraph{Flattened I3D.}
\label{architecture}
Our Flattened I3D module is designed based on the popular action recognition architecture, 3D-inflated ResNet-50 (ResNet-I3D) \cite{carreira2017quo}. ResNet-I3D can be decomposed to three parts, spatial-temporal convolutional layers $f_\textsc{conv}(\cdot)$, spatial-temporal pooling layer $f_\textsc{pool}(\cdot)$, and classification layer $f_\textsc{clf}(\cdot)$. Differing to the standard action recognition tasks, our task needs to predict not only what the action is, but also whether the frame is a start or end frame. Therefore, fine-grained temporal information is needed. Since the majority of the input frames near the action boundary might be background frames, directly conducting spatial-temporal pooling in the ResNet-I3D network may lose the action boundary information. Hence, we propose a flattened I3D, i.e., we replace $f_\textsc{pool}$ layer by $f_\textsc{flatten}$ layer that pools over spatial dimension and \textbf{concatenates} over temporal dimension. An illustrated pipeline of the Flattened I3D is in Fig.~\ref{fig:flatten_structure}. The modified architecture is as below: $$f = f_\textsc{clf} \circ f_\textsc{\textbf{flatten}} \circ f_\textsc{conv}(\cdot).$$

We optimize the Flattened I3D model with the following loss function:
\begin{small}
\begin{align}
    f^* = \argmin_{f} - \sum_{c=1}^{C+1} 1\{y_c = c\}\log(p^{c}),
\end{align}
\end{small}where $y_c$ stands for the ground truth category of the input clip, $p^c$ is the predicting score of the action start (or end) of class $c$.
During training, each batch consists of both clips with action start (or end) and background. In our case, the background can be either the clips with no actions, or the clips with actions but not action boundaries.

\paragraph{Window Based Suppression (WBS).}
\label{post-processing_with_buffer_setting}

Since frames near the real action boundary can be quite similar, a model can make multiple duplicated predictions on an action start or end.
Inspired by the Non-Maximal Suppression (NMS) in object detection, we propose a method named window based suppression (WBS), which help reduce the duplicated predictions in the online setting. Specifically,
WBS takes the prediction $\bold{p}_{k}$ from Flattened I3D module along with predictions of its prior frames in a window size $w$, i.e., $[\bold{p}_{k-w}, \bold{p}_{k-1}]$ as input, and updates the indicator vector $\bold{d}_i$ for each frame $i \in [k-w, k]$. For each action class $c$, $d_i^c$ would be set to 1 if $p_i^c$ is the highest prediction score within the window. If $p_i^c$ is not the highest prediction score, then $d_i^c$ will be set to 0. $\bold{d}_i$ is no longer to be changed once the frame $i$ has moved out of the window. WBS is illustrated in Fig.~\ref{fig:architecture}.

\section{Experiments}
\label{exp}
\begin{figure*}
    \centering
    \includegraphics[width=\textwidth]{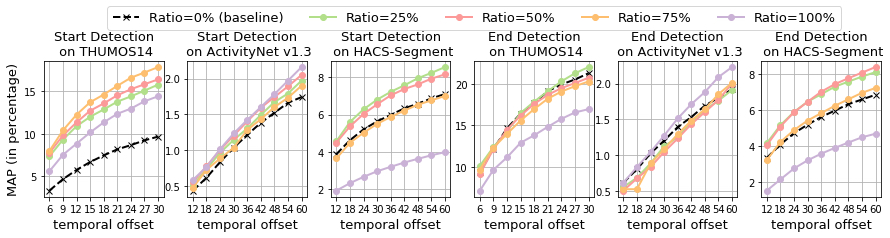}
    \caption{Performance of online action start and end detection for simple baseline classification model (ResNet-I3D) with different \emph{Ratio} in input clips on three datasets correspondingly. The baseline (without frames from time buffer) is marked as the dashed black line.}
    \label{fig:start_end_result}
\end{figure*}

\subsection{Datasets and Implementation Details}

We conduct experiments on three datasets:
\textbf{THUMOS'14}  \cite{THUMOS14} contains 1010 videos for validation and 1574 videos for testing\footnote{Since only videos in the validation and test set in THUMOS'14 contains action boundary annotations, we follow the widely applied setting to train on the validation set and evaluate on the test set.}. Videos are annotated in 20 action categories with action boundaries. \textbf{ActivityNet v1.3} contains 14,950 videos with 200 action classes\cite{caba2015activitynet} split into training and validation sets. The training set contains 10,024 videos, and the validation set contains 4,926 videos.
\textbf{HACS Segment} dataset is proposed by \cite{zhao2019hacs}. It contains 35,300 untrimmed videos over 200 classes for training and 5,530 untrimmed videos for validation.

 The Flattened I3D model takes a video clip of $l$ frames as the input, where $l$ is a configurable parameter. Details on the selection of $l$ can be referred to Sec.~\ref{ablat}. We use ResNet-I3D \cite{carreira2017quo} pre-trained on Kinetics-400 dataset as the backbone model.
 During trianing, each batch contains 128 clips. We balance the ratio between the positive and negative samples in each batch as 1:1. Standard data augmentation with random crop and random horizontal flip are used. At inference time, our model makes prediction for every 10 frames. More implementation details can be referred to the Supplementary Material.

\subsection{Evaluation Protocol and Metrics}
\label{exp:metrics}

Following the previous work~\cite{shou2018online}, we use Mean Average Precision (MAP) within the temporal offsets as the evaluation metrics in our experiments. Specifically, given a video, our approach outputs a list of tuples. Each tuple includes the timestamp of the frame, the  prediction score of action start (end), and the action class. For each positive start (end) prediction, we calculate the absolute differences between the frame index of prediction and the ground truth annotation. The prediction would be considered as correct when the difference is within the defined temporal offset (shown in Fig.~\ref{fig:illustration}), and the predicted action category is correct.

\subsection{Evaluation Results}

\subsubsection{Ablation Studies}
\label{ablat}

\paragraph{Using Frames from Buffer.}
\label{time buffer}

We first study whether future frames from the time buffer can boost the performance of the simple classification model in action start and end prediction. To evaluate it, we use the baseline model (i.e., the ResNet-I3D) with different clip inputs.
We set $k = t-\Delta$ in Fig.~\ref{fig:architecture}, i.e., all the future frames in the time buffer $\Delta$ are used as inputs to the classification model, and thus, WBS is completely turned off. We denote $\emph{Ratio} = \frac{\Delta}{l}$, which indicates that how much percentage of frames in the input clip with a length $l$ are from the buffer.
We first tune the \emph{Ratio} from $0\%$ to $100\%$ with a fixed $l=64$. The results is summarized in Fig. \ref{fig:start_end_result}. We can see that the best performance is always achieved when we include some future frames from time buffer in the input (i.e., \emph{Ratio} $> 0\%$), which proves that future frames are helpful for prediction. Comparing to the baseline (i.e., \emph{Ratio} $= 0\%$), using future frames from time buffer improves the performance by $8.2\%$, $0.4\%$, and $1.5\%$ on THUMOS'14, ActiivtyNet-v1.3, and HACS for action start detection, and by  $0.8\%$, $0.3\%$, and $1.3\%$ for the end detection, respectively.
It is interesting to observe that the best performance occurs at different \emph{Ratio} values for different datasets. This could due to the significant differences of annotations, categories and duration of actions across the datasets. In general, having a mix of preceding and succeeding frames ($0< Ratio < 100$) achieves better performance, representing that the temporal context information before and after the boundary are both important for boundary detection. Also, succeeding frames are more significant for the start detection than the end detection, which is expected, as succeeding frames of action start contains the action, while those of action end contains only background.

\begin{figure}
    \centering
    \begin{minipage}{.4\textwidth}
        \centering
    \centering
    \includegraphics[width=1\textwidth]{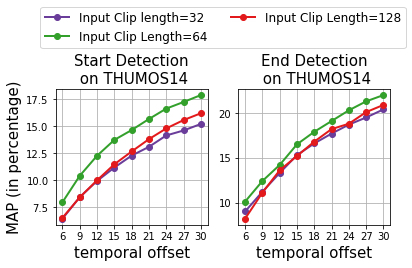}
    \caption{Results of different input clip length $l$ to the ResNet-I3D model on THUMOS'14 dataset.}
    \label{fig:changing_input_size}
    \end{minipage}
    \hspace{10pt}
    \begin{minipage}{0.4\textwidth}
    \centering
    \includegraphics[width=1\textwidth]{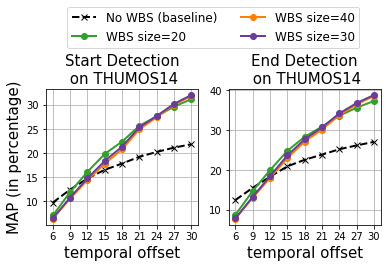}
    \caption{Results of different window sizes in WBS on THUMOS'14.}
    \label{fig:Window_based_suppression}
    \end{minipage}
\end{figure}

We then study the performance of different input video clip size $l$. We fix \emph{Ratio}$=75\%$. Fig. \ref{fig:changing_input_size} shows the results on THUMOS'14, with $l$ equals to 32, 64, and 128 frames, respectively. From the figure, the best results is at $l=64$ for both action start and end detection. We also observe the similar trend on ActivityNet-v1.3 and HACS datasets. In the following experiments, we fix the input size at $l=64$ frames.

\paragraph{Flattened I3D.}

\begin{figure*}
    \centering
    \includegraphics[width=\textwidth]{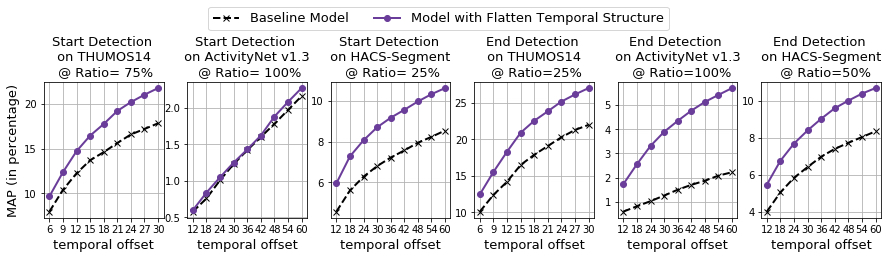}
    \caption{Comparison of model with flattened temporal structure (Flattened-I3D) and baseline model (ResNet-I3D) on action start and end detection.}
    \label{fig:flatten_feature_start_end}
\end{figure*}

We then compare our proposed Flattened-I3D model with the baseline ResNet-I3D architecture. We still turn off the WBS for comparison. We select the best ratio setting based on Fig.~\ref{fig:start_end_result}, i.e., \emph{Ratio} equals to $75\%, 100\%, 25\%$ for start detection and $25\%, 100\%, 50\%$ for end detection on THUMOS'14, ActivityNet-v1.3, and HACS-Segment dataset, respectively. The results are summarized in Fig. \ref{fig:flatten_feature_start_end}. We can observe that Flattened-I3D surpasses ResNet-I3D on all the cases, demonstrating that fine-grained temporal structures provide additional benefits to the action boundary detection.

\paragraph{Window Based Suppression.}

Next, we evaluate effectiveness of the WBS. Fig.~\ref{fig:Window_based_suppression} shows the action start and end detection performance on THUMOS'14 using Flattened I3D with and without WBS. From the figure, we can see that with the WBS, the performance improves by up to $10.2\%$ and $11.8\%$ on detection of action start and end, respectively.
We also compare results of using WBS with different window sizes, i.e., $w = k - (t-\Delta)$ in Fig. \ref{fig:Window_based_suppression}. It is interesting to see that though $w=30$ frames gives the best performance, the performance overall is not very sensitive to the window size, as long as WBS is used. We set the window size to $w=30$ frames in the rest of the experiments.

\subsubsection{Quality Analysis on Datasets}
\label{sec:QAD}

\begin{table}[]
    \centering
    \small
    \begin{tabular}{c|c|c}
         &  \multicolumn{2}{c}{Mean Annotation Error Rate}  \\
         Datasets & action start & action end\\
    \hline
    \hline
        THUMOS'14 &          1.0\%   & 0.3\%\\
        ActivityNet-v1.3 &   5.9\%   & 4.2\%\\
        HACS-Segment &       0.9\%   & 2.9\%\\
    \end{tabular}
    \vspace{3pt}
    \caption{The mean of annotation error rates over all categories, on three datasets.}
    \label{tab:verify_as}
\end{table}

THUMOS'14, and ActivityNet-v1.3 are used in developing the online start and end detection algorithms today~\cite{shou2018online,gao2019startnet}. However, there lacks of careful studies on the quality of the action boundary annotations in these datasets. Poor annotation qualities might lead results to be fragile. To make the study more solid,
we verify action start and end annotation quality on all three datasets using Amazon Mechanical Turk. Specifically, for each frame marked as action start (end) in the video, we crop its nearby 20 frames to form a snapshot and ask Turkers whether the snapshot contains an action boundary. If the snapshot only contains background or action, but not an action start or end,  it would be labeled as an ``Error''. Each snapshot is verified by 5 Turkers. We summarize the human verification results for three datasets in Table \ref{tab:verify_as}. We can see that both THUMOS'14 and HACS-segment datasets have high qualify in their annotations for action boundaries, while the quality of the ActivityNet-v1.3 is relatively poor. It's worth to mention that the annotation error rates for some classes in the ActivityNet are above 20-30\%. Please see Suppl. Material for more details.

\subsubsection{Results Comparison}
\begin{table*}[]
    \centering
    \resizebox{\linewidth}{!} {
    \begin{tabular}{c||c||c|c|c|c|c|c|c|c|c|c }
          Task & Offsets (frames)              & 30      & 60     & 90  & 120     & 150 & 180 & 210 & 240 & 270 & 300    \\
         \hline
         & SceneDetect~\cite{SceneDetect}  & 1.0 & 2.0 & 2.3 & 3.1 & 3.6 & 4.1 & 4.7 & 5.0 & 5.1 & 5.2\\
         & ShotDetect~\cite{ShotDetect}   & 1.1 & 1.9 & 2.3 & 3.0 & 3.4 & 3.9 & 4.3 & 4.5 & 4.6 & 4.9\\
          & ODAS \cite{shou2018online}                   & 3.1    & 4.3   & 4.7  & 5.4 & 5.8 & 6.1 & 6.5 & 7.2 & 7.6 & 8.2\\
           & StartNet (C3D+LocNet) \cite{gao2019startnet}  & 6.8    & 8.0   & 9.4 & 10.1 & 10.6 & 10.9 & 10.9 & 11.1 & 11.2 & 11.2 \\
         Detect & StartNet (LSTM+LocNet) \cite{gao2019startnet} & 19.5   & 27.2  & 30.8 & 33.9 & 36.5 & 37.5 & 38.3 & 38.8 & 39.5 & 39.8  \\
         \cline{2-12}
         Start& Flattened I3D w/o time buffer & 8.3  & 11.6  & 14.1 &  16.6 & 18.4 & 19.9 & 21.5 & 23.1 & 24.1 & 25.3\\
         &\textbf{BF-I3D}    & \textbf{32.0} & \textbf{41.7}  & \textbf{45.3} & \textbf{47.9}& \textbf{50.3}& \textbf{51.3}& \textbf{51.9}& \textbf{52.5}& \textbf{53.2} & \textbf{53.6}\\
         \hline
         Detect & Flattened I3D w/o time buffer & 25.4& 31.3& 34.1 & 36.5 & 38.2 & 40.6 & 42.8 & 44.0 & 45.2 & 46.5\\
         End &\textbf{BF-I3D}    & \textbf{38.8} & \textbf{49.1} & \textbf{52.7} & \textbf{53.9} & \textbf{55.2} & \textbf{55.6} & \textbf{56.6} & \textbf{57.2} & \textbf{57.8} & \textbf{58.3}\\
    \end{tabular}
    }
    \vspace{1pt}
    \caption{Results comparison on THUMOS14 for action start and end detection.}

    \label{tab:SOTA_TH14}
\end{table*}

We compare our BF-I3D solution to the baseline solution (Flattened I3D without time buffer) as well as the state-of-the-art (SoTA) online detection approaches. We use $Ratio= 75\%$, clip length $l=64$, and WBS window size $w=30$.  Table~\ref{tab:SOTA_TH14} summarizes the results on the THUMOS'14 dataset under temporal offsets from 30 frames to 300 frames. SceneDetect~\cite{SceneDetect} and ShotDetect~\cite{ShotDetect} are both two-stage methods to first detect the change boundaries in the video and then classify the actions\footnote{The numbers are obtained from the paper~\cite{gao2019startnet}.}. ODAS~\cite{shou2018online} is a sliding window based C3D~\cite{tran2015learning} model. StartNet~\cite{gao2019startnet} is an LSTM based approach that is optimized with Policy Gradient~\cite{sutton2018reinforcement}. From the table, we can see that, for both action start and end detection, BF-I3D significantly outperforms the Flattened I3D without time buffer, demonstrating the effectiveness of using the time buffer. Moreover, BF-I3D outperforms all the other approaches in all different settings with a large margin. It surpasses the SoTA StartNet by $13.8\%$ on offset of $300$ frames. It is worth to note that the overall size of time buffer is $\text{Ratio} \times l + w = 78$, which is only about 2-3 seconds time delay. Similar results are observed on ActivityNet and HACS datasets. Please refer to the Suppl. Material for more details.

As our model observes 2-3 seconds additional frames from the future, we are also interested in comparing the performance of our solution to the others under the setting that the model observes the same number of frames during evaluation. To achieve this, we take the time buffer into account and re-align the temporal offsets in comparison in Table~\ref{tab:SOTA_TH14_model}, e.g., we compare BF-I3D at offsets=30 to other approaches (without time buffer) at offset=120 to keep 90 frames as the room for the buffer.
We would like to point out that such comparison is fair. When evaluating SoTA approaches (without time buffer) at offset=120, any correct predictions under offset of 120 frames would be counted as positives. In this way, SoTA approaches are able to utilize additional 90 frames as inputs to their systems to make final predictions, which is equivalent to a 90-frame buffer in BF-I3D.
Results show that under this setting, our approach can still surpass the SoTA by a margin when the offset goes beyond $150$ frames.

\begin{table*}[]
    \centering
    \resizebox{.85\linewidth}{!} {
    \begin{tabular}{c||c||c|c|c|c|c|c|c }
          Task & Offsets (frames)              & 120     & 150 & 180 & 210 & 240 & 270 & 300 \\
         \hline
          & ODAS \cite{shou2018online}                   & 5.4 & 5.8 & 6.1 & 6.5 & 7.2 & 7.6 & 8.2\\
         Detect & StartNet (C3D+LocNet) \cite{gao2019startnet}  &  10.1 & 10.6 & 10.9 & 10.9 & 11.1 & 11.2 & 11.2 \\
         Start & StartNet (LSTM+LocNet) \cite{gao2019startnet} & 33.9 & 36.5 & 37.5 & 38.3 & 38.8 & 39.5 & 39.8  \\
         \cline{2-9}
         & \textbf{BF-I3D}    & 32.0 & \textbf{41.7}  & \textbf{45.3} &
         \textbf{47.9}& \textbf{50.3}& \textbf{51.3}& \textbf{51.9}\\
    \end{tabular}
    }
    \vspace{5pt}
    \caption{Results comparison on THUMOS14 for action start detection after the offset re-alignment by taking the time buffer into account. In each column, all the models in the table now observe the same number of frames.}
    \label{tab:SOTA_TH14_model}
\end{table*}

\vspace{-0.1in}
\section{Conclusion}
\label{conclusion}

In this paper, we have adopted a new problem setting for the online temporal action detection in live streaming videos by taking the time delays into account. The setting is practical, and results on multiple benchmarks have shown that leverage the tiny near-term future information from the time delays can significantly improve the performance of online action start and end detection. While we only propose a simple new paradigm, i.e., the BF-I3D solution in this paper, one future direction can be investigating an attention based architecture that can leverage the information in the time buffer in an optimized way.

\appendix

\section{Appendix}

In this supplementary material, we provide more results and implementation details that are skipped in the main paper due to the limit of space. First, we present the implementation details in Section~\ref{impl}. The details of quality analysis on multiple datasets are introduced in Section~\ref{qad}. More results and analysis of the proposed \emph{buffer based flattened I3D} (\emph{BF-I3D}) are in Section~\ref{supp_exp}.

\subsection{Implementation Details}
\label{impl}
In this section, we provide the training details that is skipped in the main paper. Specifically, the standard data augmentation with random crop and random horizontal flip are used during training.  For each video clip, we first resize the spatial resolution of videos to $340 \times 256$, randomly crop a $224 \times 224$ patch, and then randomly flip the patch horizontally to form an input to our model. We summarize the hyper-parameter for three datasets as below:

\textbf{THUMOS'14}: We follow the author's original setting~\cite{shou2018online} to train on the validation set and evaluate on the test set. We train our model for 300 epochs with learning rate starting from 0.01 and decaying to 0.001 and 0.0001 at 100 and 200 epochs, respectively.

\textbf{ActivityNet v1.3}: We train our model on training set and evaluate on the validation set. Our model is trained for 150 epochs. The learning rate is intialized as 0.001 and is dropped to 0.0001 and 0.00001 at 50 and 100 epochs, respectively.

\textbf{HACS Segment}: Our model is trained for 15 epochs on the training split. The learning rate is initalized as 0.01 and is dropped to 0.001 and 0.0001 at 5 and 10 epochs, respectively.

\subsection{Quality Analysis on Datasets}
\label{qad}
\begin{figure}[hbt!]
    \centering
    \includegraphics[width=\textwidth]{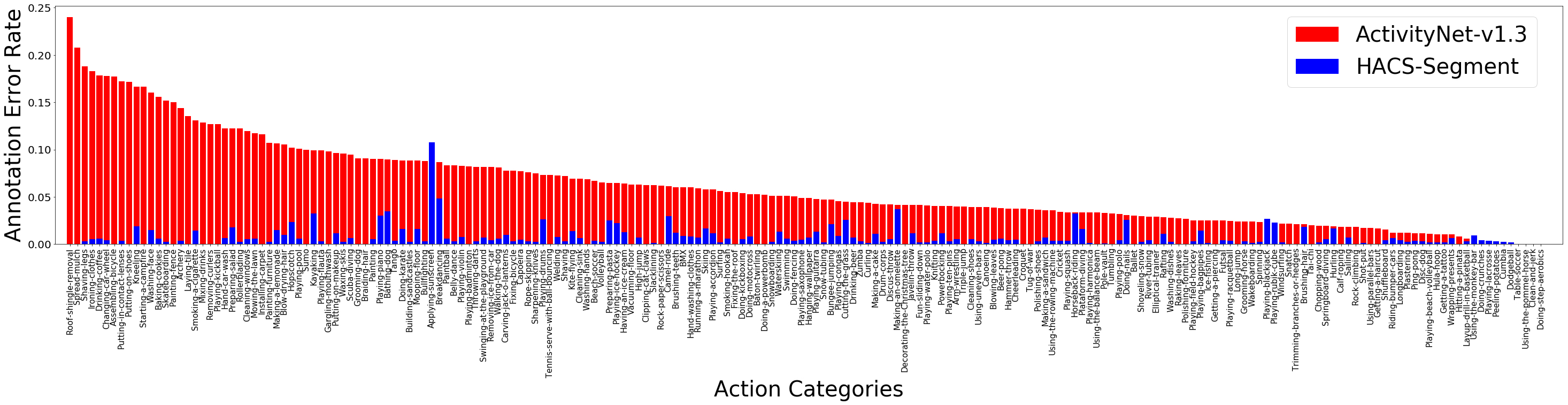}

    \caption{Per class annotation error rate (normalized to 1) for action start on ActivityNet v1.3 and HACS dataset. }

    \label{fig:per_class_incorrect_labeled_rate}
\end{figure}

We summarize the per class annotation error rate for action start on ActivityNet v1.3 and HACS-segment dataset in Fig. \ref{fig:per_class_incorrect_labeled_rate}. The red bar represents the annotation error rate for ActivityNet dataset, while the blue bar stands for the annotation error rate for HACS dataset. It could be observed that in general the error rates for HACS dataset are much lower than those of ActivityNet dataset. On ActivityNet v1.3, the annotation error rates for some categories, \emph{e.g.} Roof Shingle Removal and Spread Mulch, are over $20\%$.

\subsection{Additional Evaluation Results}
\label{supp_exp}
\subsubsection{Results on HACS and ActivityNet Datasets}

Results of action start and end detection on HACS and ActivityNet-v1.3 datasets are in Table~\ref{tab:SOTA_HACS} and Table~\ref{tab:SOTA_ANET}, respectively. We are the first to conduct online detection experiments on the HACS dataset, so there is no literature work to compare. In Table~\ref{tab:SOTA_HACS} we mainly compare our BF-I3D solution to the baseline (Flattened I3D) without using time buffer. In BF-I3D, we use $Ratio= 25\%$, clip length $l=64$, and WBS window size $w=30$, so the overall size of time buffer is $\text{Ratio} \times l + w = 46$ frames, which is only about 2 seconds time delay. Results on the HACS dataset under temporal offsets from 1 sec to 10 secs (i.e. from 30 frames to 300 frames) are reported. We can see from the results that under all the temporal offset settings, for both action start and end detection, BF-I3D significantly outperforms the Flattened I3D without time buffer, demonstrating the effectiveness of using the time buffer.

Table~\ref{tab:SOTA_ANET} summarizes the results on ActivityNet-v1.3 dataset. We follow the literature to compare the results with the temporal offset at 300 frames only~\cite{shou2018online}. In BF-I3D, we use $Ratio= 100\%$, clip length $l=64$, and WBS window size $w=30$, so the overall size of time buffer is $\text{Ratio} \times l + w = 94$ frames, which is only about 3 seconds time delay. We can see that BF-I3D outperforms the Flattened I3D without time buffer, demonstrating again the effectiveness of using the time buffer. Moreover, our approach can surpass all the solutions except for the StartNet. One potential reason is the long duration of action instances in ActivityNet dataset might benefit more on LSTM based approach (StartNet). To better understand the results, we form a subset of ActivityNet, which only contains videos with action instance that is shorter than $30\%$ of the video length and re-evaluate our approach on this subset. Our BF-I3D can achieve performance at $\textbf{19.4\%}$ with the offset at 300 frames, outperforming the StartNet. Also, while StartNet is using carefully fine-tuned features pre-trained on large-scale datasets (i.e., Kinetics or ImageNet), and ODAS is using well fine-tuned features extracted from TSN~\cite{wang2016temporal}, we simply train our own feature on ActivityNet without any bells and whistles. In addition, as reported in Table 1 in the main paper, the high annotation error rate on ActivityNet-v1.3 dataset may affect the model performance.  % It should also be noticed that StartNet leverages additional input (optical flow), while in our model we only use RGB as the input.

\begin{table*}[]
    \centering
    \resizebox{\linewidth}{!} {
    \begin{tabular}{c||c||c|c|c|c|c|c|c|c|c|c }
          Task & Offsets (frames)              & 30      & 60     & 90  & 120     & 150 & 180 & 210 & 240 & 270 & 300    \\
         \hline
         Detect & Flattened I3D w/o time buffer & 7.1 & 8.9 & 10.2 & 11.2 & 12.3 & 13.4 & 14.3 & 15.2 & 16.1 & 16.9 \\
         Start & \textbf{BF-I3D}    & \textbf{12.5} & \textbf{16.9} & \textbf{19.3} & \textbf{21.2} & \textbf{22.6} & \textbf{24.0} & \textbf{25.3} & \textbf{26.3} & \textbf{27.3} & \textbf{28.3}\\
         \hline
         Detect & Flattened I3D w/o time buffer & 6.1 & 7.9 & 9.0 & 10.0 & 10.8 & 11.7 & 12.4 & 13.2& 14.0 & 14.7 \\
         End &\textbf{BF-I3D}      & \textbf{11.6} & \textbf{16.4} & \textbf{18.9} & \textbf{20.6} & \textbf{22.1} & \textbf{23.6} & \textbf{24.9} & \textbf{26.0} & \textbf{27.0}& \textbf{27.9}\\

    \end{tabular}
    }
    \vspace{0.5pt}
    \caption{Results comparison on HACS-Segment for action start and end detection.}
    \label{tab:SOTA_HACS}
\end{table*}

\begin{table}[]
    \centering
    \begin{tabular}{c||c||c }
          Task & Offsets (frames)             & 300    \\
         \hline
          & SceneDetect \cite{SceneDetect} & 4.7\\
          & ShotDetect \cite{ShotDetect} & 6.1 \\
          & ODAS \cite{shou2018online}                  & 8.3  \\
         Detect & StartNet \cite{gao2019startnet}   & 13.5 \\
         \cline{2-3}
         Start& Flattened I3D w/o time buffer & 7.2 \\
         &\textbf{BF-I3D}    & \textbf{9.9} \\
         \hline
         Detect & Flattened I3D w/o time buffer & 7.1\\
         End &\textbf{BF-I3D}      & \textbf{10.4} \\

    \end{tabular}
    \vspace{3pt}
    \caption{Results comparison on ActivityNet-v1.3 for action start and end detection.}
    \label{tab:SOTA_ANET}
\end{table}

\bibliography{main}

\end{document}